# Conversational No-code, Multi-agentic Disease Module Identification and Drug Repurposing Prediction with ChatDRex

Simon Süwer[1*], Kester Bagemihl[1*], Sylvie Baier[1,2], Lucia Dicunta[1], Markus List[2,3], Jan Baumbach[1,4,†], Andreas Maier[1,†], Fernando M. Delgado-Chaves[1,†]

[*] These authors would like to be regarded as shared first authors of the paper

[†] These authors would like to be regarded as shared last authors of the paper

[1] Institute for Computational Systems Biomedicine, University of Hamburg, Hamburg, Germany

[2] Data Science in Systems Biology, TUM School of Life Sciences, Technical University of Munich, Freising, Germany

[3] Munich Data Science Institute, Technical University of Munich, Garching, Germany

[4] Institute for Mathematics and Computer Science, University of Southern Denmark, Odense, Denmark

## Abstract

Repurposing approved drugs offers a time-efficient and cost-effective alternative to traditional drug development. However, *in silico* prediction of repurposing candidates is challenging and requires the effective collaboration of specialists in various fields, including pharmacology, medicine, biology, and bioinformatics. Fragmented, specialized algorithms and tools often address only narrow aspects of the overall problem. Heterogeneous, unstructured data landscapes require the expertise of specialized users. Hence, these data services do not integrate smoothly across workflows.

With ChatDRex, we present a conversation-based, multi-agent system that facilitates the execution of complex bioinformatic analyses aiming for network-based drug repurposing prediction. It builds on the integrated systems medicine knowledge graph (NeDRex KG). ChatDRex provides natural language access to its extensive biomedical knowledge base. It integrates bioinformatics agents for network analysis, literature mining, and drug repurposing. These are complemented by agents that evaluate functional coherence for in silico validation. Its flexible multi-agent design assigns specific tasks to specialized agents, including query routing, data retrieval, algorithm execution, and result visualization. A dedicated reasoning module keeps the user in the loop and allows for hallucination detection.

By enabling physicians and researchers without computer science expertise to control complex analyses with natural language, ChatDRex democratizes access to bioinformatics as an important resource for drug repurposing. It enables clinical experts to generate hypotheses and explore drug repurposing opportunities, ultimately accelerating the discovery of novel therapies and advancing personalized medicine and translational research. ChatDRex is publicly available at apps.cosy.bio/chatdrex.

## Introduction

The development of novel pharmaceuticals requires substantial time and financial investments. Additionally, it is often accompanied by a high risk of failure in the late stages of clinical trials [1–3]. To address these challenges, there is an increasing emphasis on repurposing already formulated drugs for novel therapeutic applications. This approach leverages comprehensive pharmacological, clinical, or genomic data to identify potential new indications for existing drugs [1,2]. This approach not only significantly accelerates the development process but also reduces the risk of clinical failures, as the safety and tolerability of the substances are already established [1,2].

In particular, systems medicine provides a conceptual framework for understanding disease as a complex, multi-scale process that arises from interactions across molecular, cellular, and clinical levels [4].
Network medicine operationalizes this paradigm by translating system-level views into explicit computational models, which are typically represented as biological and pharmacological networks [5]. Researchers have found these network-based methodologies to be helpful in identifying promising drug repurposing candidates, thereby speeding up the transition from discovery to clinical application [6,7]. Since networks capture the relationships between genes, drugs, proteins, and diseases, they enable the identification of novel indications for existing pharmaceutical agents [8]. This type of network-based prediction of drug repurposing candidates has already been demonstrated in the context of Alzheimer's disease and colorectal cancer [9–12].

A generic workflow for network-driven drug repurposing was implemented and demonstrated within the NeDRex (Network-based Drug Repurposing and Exploration) project [7]. Central to this methodology are concepts such as *seed genes*, which serve as the starting point for drug repurposing queries, typically genes known to be implicated in a specific disease. *Disease modules* are clusters of interconnected genes driving to the manifestation of a disease phenotype. For example, in Huntington's disease, the *HTT* gene is the main disease-causing gene and the associated network consisting of genes involved in neuronal damage forms the Huntington's disease module. *Disease module identification* algorithms, such as DIAMOnD (DIseAse MOdule Detection), are used to extract these modules from complex biological networks [13]. Once a disease module is identified, *drug prioritization* algorithms rank drugs that target the identified module, thereby allowing researchers to focus on the most promising candidates for repurposing [14]. These algorithms utilize network topology, often through proximity measures, to determine the association between a drug target and the disease module, providing insights into the drug's potential efficacy [14].

Nevertheless, the field faces considerable challenges due to the substantial and intricate nature of biomedical information that must be examined to identify drug candidates with potential for novel indications. The NeDRex platform, primarily composed of the NeDRex KG and the NeDRex API, was developed to address the identified challenges by providing a comprehensive suite of tools for network-based drug repurposing [7]. Utilizing these advanced tools requires a decent degree of bioinformatics expertise or dedicated training, which poses a significant hurdle for many clinicians despite their valuable pharmacological or disease-specific expertise [15]. Consequently, there is a pressing need for more accessible and user-friendly interfaces to bridge the gap between complex bioinformatics tools and the researchers and clinicians who could benefit most from their insights, ultimately accelerating the development of new treatments and personalized medicine approaches [16,17].

Large language models (LLMs) represent a transformative advance in this regard, offering unprecedented capabilities in natural language processing, including question answering and even medical reasoning [18,19]. The success of these models can be attributed to their extensive training on vast quantities of text data, which has enabled them to comprehend and generate coherent natural language [19]. These models have revolutionized biomedical data analysis by providing powerful text-understanding capabilities [20–22]. However, their generalized nature often limits their effectiveness in specialized biomedical domains [23,24]. Additionally, it has been observed that LLMs tend to generate hallucinations, producing plausible yet erroneous or fallacious information [25,26]. This phenomenon poses a significant challenge in high-risk domains such as biomedicine. The unreliability of LLMs underscores the necessity for domain-specific adaptations and rigorous validation when employing LLMs in biomedical research and clinical decision-making processes [25,26]. To overcome this limitation, Retrieval-Augmented Generation (RAG) and In-Context Learning (ICL) techniques promise to enhance LLMs by incorporating external knowledge retrieval, thereby improving the accuracy and relevance of responses in specialized fields [19,27]. LLMs can further be organized as agents, which are autonomous entities designed to perform specific tasks. Each agent operates without supervision, allowing them to pursue individual objectives and implement different strategies, including analyzing external data [28]. To execute this task, LLM agents possess a memory that enables them to acquire knowledge from prior interactions and to preserve context over extended periods of time [28]. In particular in the area of drug repurposing, LLM agents can work towards the goal of providing accurate biomedical insights by autonomously accessing data services, retrieving relevant information and iteratively refining their responses based on user feedback [29]. Through inter-agent communication, these specialized agents interact with databases, execute algorithms, and generate human-understandable explanations. This collaborative approach bridges the gap between complex computational tools and the needs of researchers and clinicians.

Building on these advancements in LLMs and agent technology, we introduce ChatDRex[1], an intuitive conversational agent designed to assist with identifying disease modules and repurposing drugs. ChatDRex leverages a multi-agent architecture to provide a user-friendly interface to essential functions of the NeDRex platform, automating complex analyses and enhancing result visualization (Fig. 1). The objective of ChatDRex is to democratize access to computational drug repurposing tools and facilitate the identification of potential treatment options for existing diseases by both researchers and clinicians. Additionally, ChatDRex incorporates network-driven drug repurposing strategies into an interactive, conversation-based system. This system is designed to help researchers and clinicians identify potential treatment options for existing diseases.

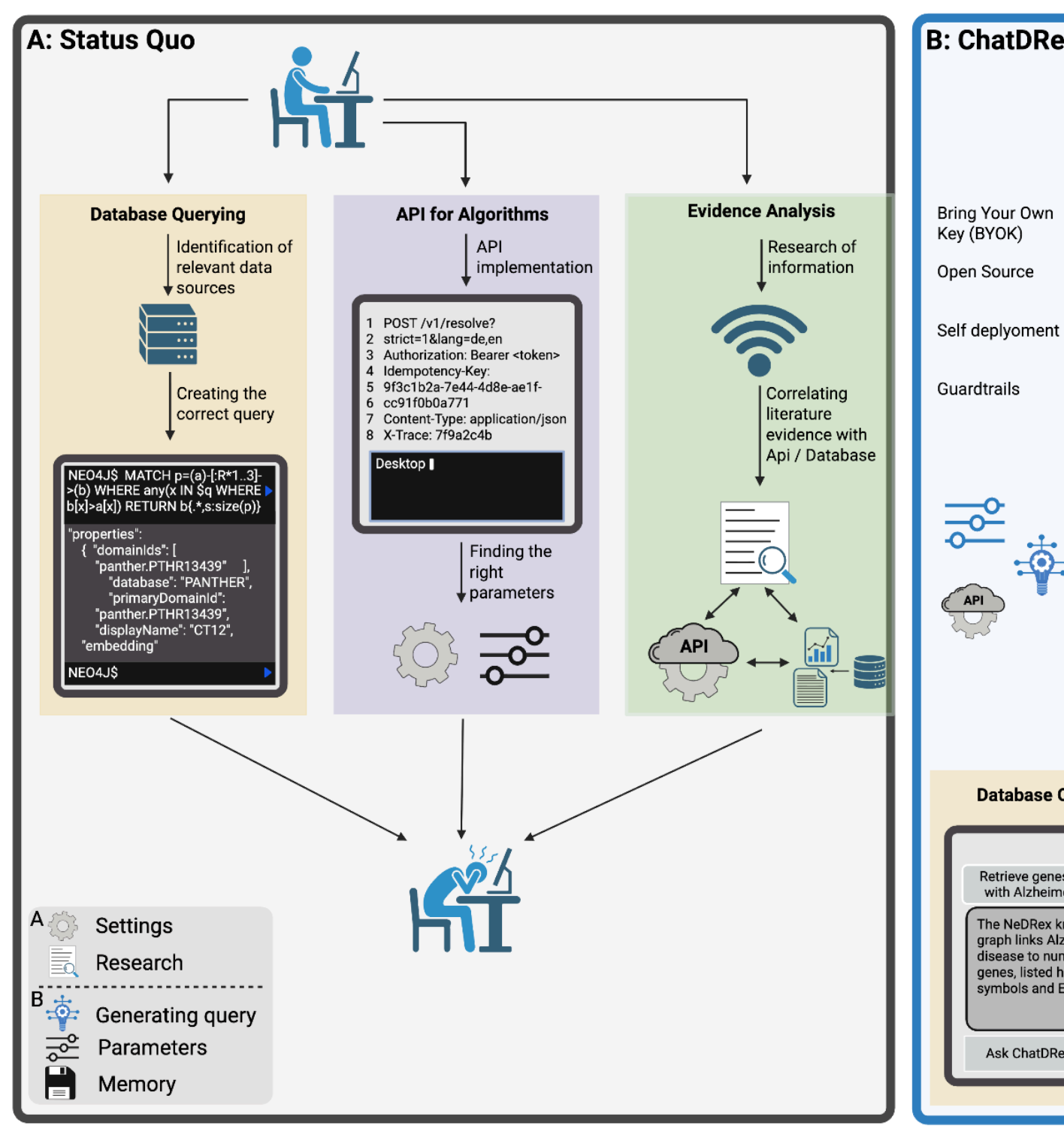

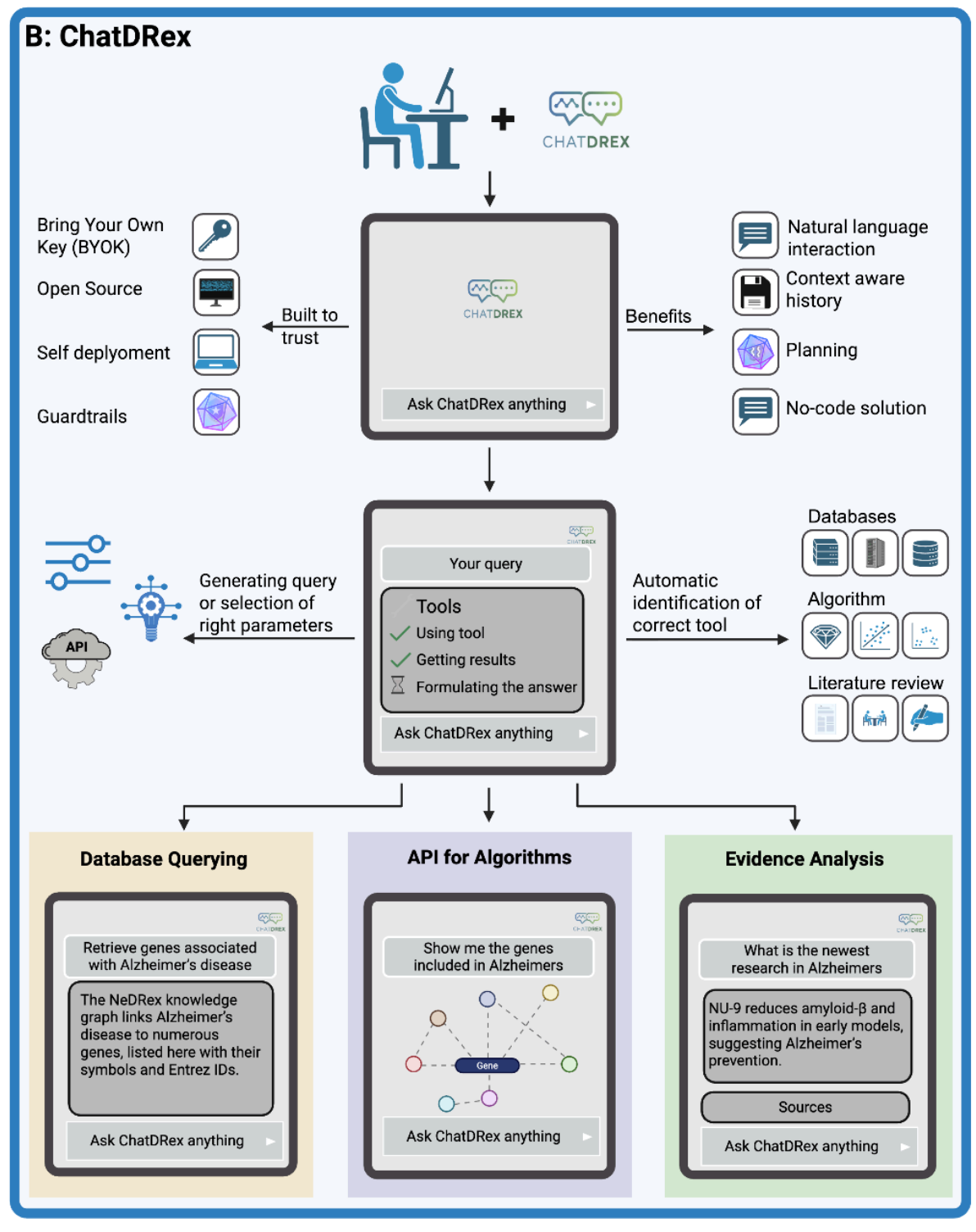

[1] https://apps.cosy.bio/chatdrex

**Figure 1**: **ChatDRex Overview:** The figure compares the traditional manual workflow (A) for biomedical analyses with the ChatDRex approach (B). Presently, users are required to formulate database queries, configure algorithmic APIs, and manually correlate the results with literature references. ChatDRex integrates these steps into a single, natural language-based workflow. ChatDRex automatically selects tools and generates the appropriate parameters. This capability allows ChatDRex to integrate relevant information from databases, algorithms, and literature into a cohesive result.

## Methods

ChatDRex streamlines complex, network-based drug repurposing analyses with a conversational interface. This approach is designed to address potential biases and hallucinations in LLMs by integrating directly with validated biomedical databases and specialized bioinformatics tools. It ensures data accuracy and enables sophisticated analyses without requiring specialized computational expertise.

The implemented multi-agent architecture is inspired by the Chain-of-Thought (CoT) paradigm [30], but adapted to the modularity of complex systems. In contrast to the original CoT, in which a single model sequentially externalizes the intermediate steps of reasoning [31]. Our approach delegates reasoning to a set of specialized agents, each of which has a clearly defined role in the drug repurposing workflow (Fig. 2). A central planning agent functions as a meta-controller, breaking down user requests into subtasks, iteratively assigning them to the appropriate expert agents, and merging their results into a coherent outcome [32]. Consequently, the architecture enables task planning and orchestration, allowing each agent to focus on a small subtask using CoT [31].

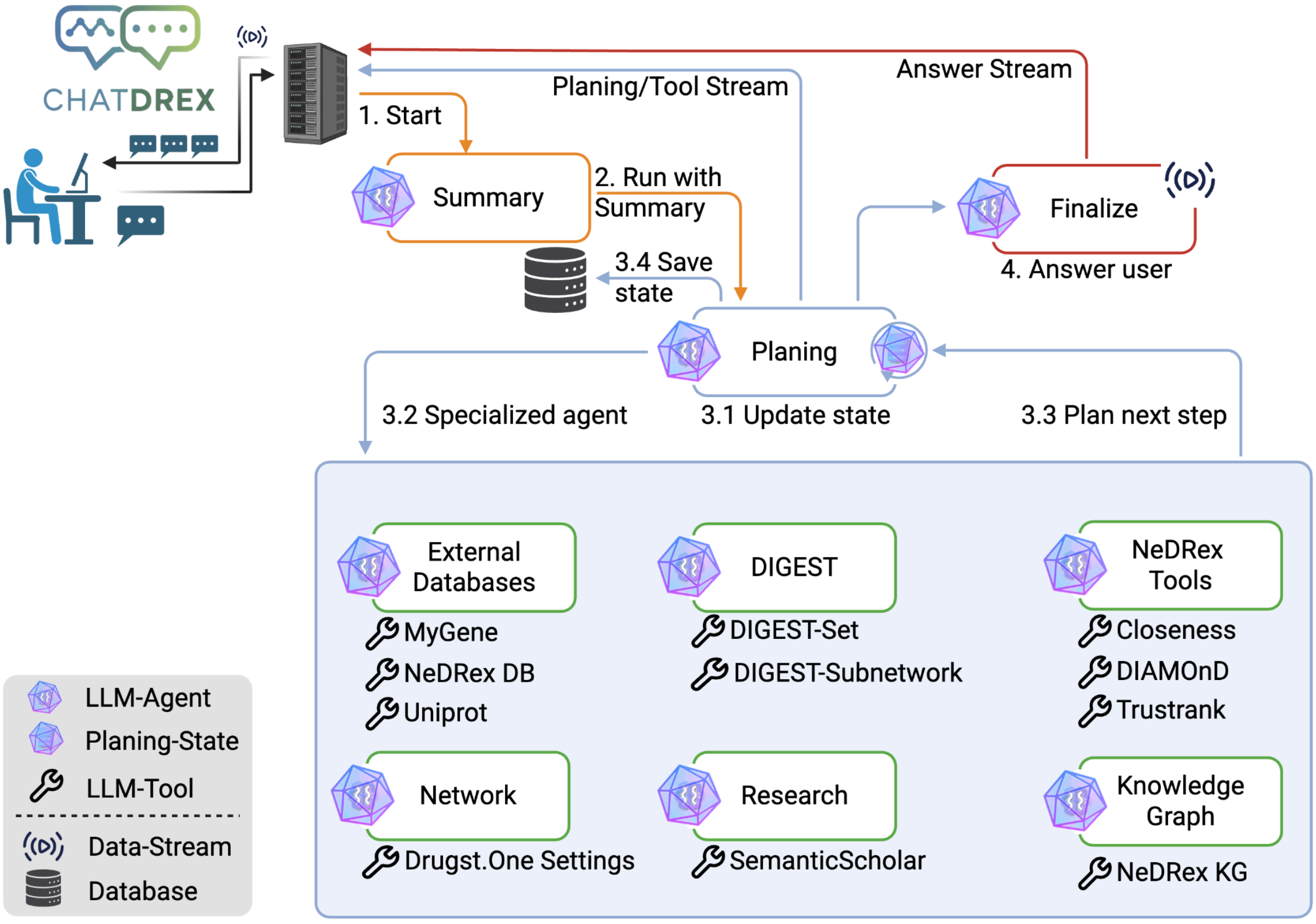

**Figure 2**: **ChatDRex Multi-Agent architecture:** The system is orchestrated by a central planning agent that routes user queries to specialized agents according to task type. The DIGEST agent performs functional coherence and enrichment analyses [33]. Our NeDRex

agents provide access to the NeDRex KG and enable network-based drug repurposing analyses via the NeDRex API [7]. The Research agent conducts literature retrieval, while an external database agent enriches queries with additional external knowledge sources. A Network agent controls and adapts the visualization of interaction networks using a Drugs.One-powered graph interface [6]. Finally, the Finalize agent integrates and synthesizes the outputs of all agents into a coherent response.

## ChatDRex Agent Architecture

ChatDRex is a multi-agent system (Fig. 2), where all agents are constructed based on the same architectural principles (Fig. 3). Each agent integrates a unified control loop consisting of few-shot prompting, which is part of ICL, short-term and long-term memory access, internal planning routines, and external tool integration. Our lightweight Summary agent maintains contextual continuity by compressing prior interaction states (S. 2.1). This reduces token usage and enables controlled memory resets. Prompt-injection safeguards are applied at this stage, which is particularly relevant given that LLMs are systematically vulnerable to jailbreaks and instruction-overriding attacks [34–36]. According to the available summary and the new user prompt, the coordination of tasks is the responsibility of a Planning agent. This agent employs an LLM-based decision process to dynamically direct user input to the most suitable specialized agents (S. 2.2). The selection of agents is formulated as a multi-step problem, ensuring efficient task decomposition and specialization [32]. If biological validation or interpretation is requested by the user, this is provided by the DIGEST agent (S. 2.3). It uses DIGEST to evaluate the functional coherence of gene sets and disease-associated subnetworks. Conventional enrichment approaches, which enumerate overrepresented terms, are contrasted with DIGEST, which quantifies internal functional consistency through comparison with randomized reference sets. This process provides empirical significance estimates [29]. Within ChatDRex, DIGEST is employed to assess the biological plausibility of disease modules derived from network analyses, as well as user-provided gene sets in the absence of network context.

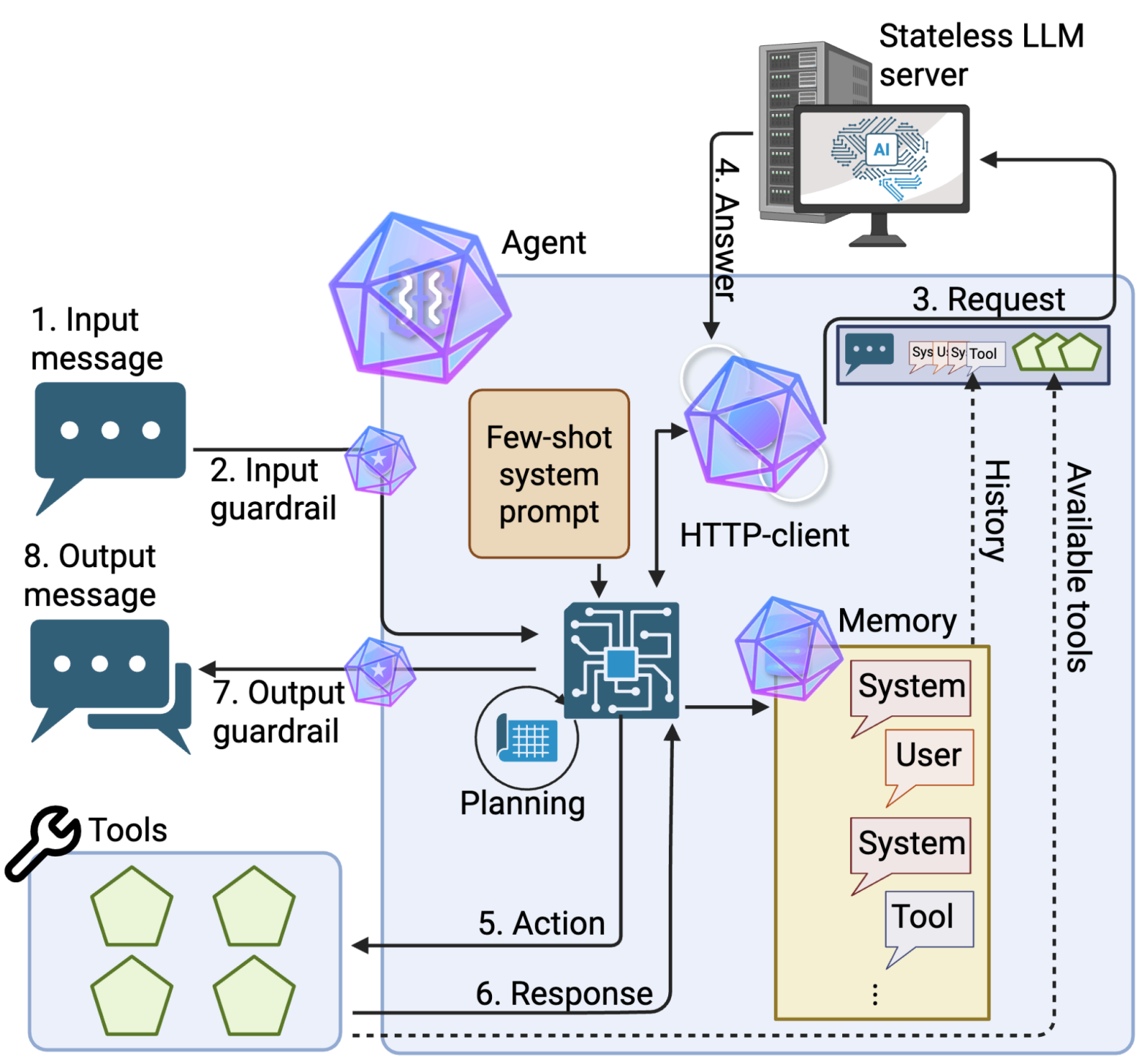

**Figure 3**: **ChatDRex agent architecture:** (1) The incoming user messages are enriched by a few-shot system prompt together with relevant memory contents, thereby documenting previous interactions between user, system, and invoked tools in a structured form. This context package is then transmitted to a stateless LLM server (open source hosted, or proprietary (3)), which processes the request independently of prior states. The server's response is subsequently evaluated by an interpretation module that decides whether immediate feedback can be returned to the user or whether additional tools need to be engaged [37,38] (4-8). When tools are invoked, their outputs are written back into the respective agent's memory (5-6), thus closing the CoT [31] (Planning). In addition, ChatDRex can incorporate optional safeguard layers. Input guardrails mitigate prompt injection risks before queries reach the LLM (2), while output guardrails constrain generated responses (6) to domain-specific requirements and filter potentially hallucinatory content (7) [39]. These safeguards are not fixed components of every agent but can be activated depending on the task. This selective integration allows the system to balance robustness against computational efficiency in a fine-grained manner.

Network-driven analyses are conducted by the NeDRex agent, which provides algorithmic access to the integrated NeDRex network database (S. 2.4). This agent orchestrates established network-medicine tools, including DIAMOnD for disease module expansion from seed genes [14]. TrustRank is a methodology that facilitates the prioritization of drug targets based on their proximity to trusted disease nodes [40]. While Closeness Centrality serves to rank candidates according to their global network distances [7]. The integration of these tools facilitates the systematic identification of disease-relevant subnetworks and candidate therapeutics within a unified interaction graph, thereby constituting the computational core of ChatDRex's network-based drug repurposing capabilities. The Knowledge Graph agent is responsible for the structured knowledge retrieval process (S. 2.5). This agent translates natural-language queries into schema-constrained graph queries over the NeDRex KG (Fig. 4). The agent enhances its resilience, traceability, and resistance to hallucinations by integrating semantic query decomposition, embedding-based node matching, and deterministic Cypher query generation. Additionally, it employs fallback strategies that are guided by graph-based retrieval.

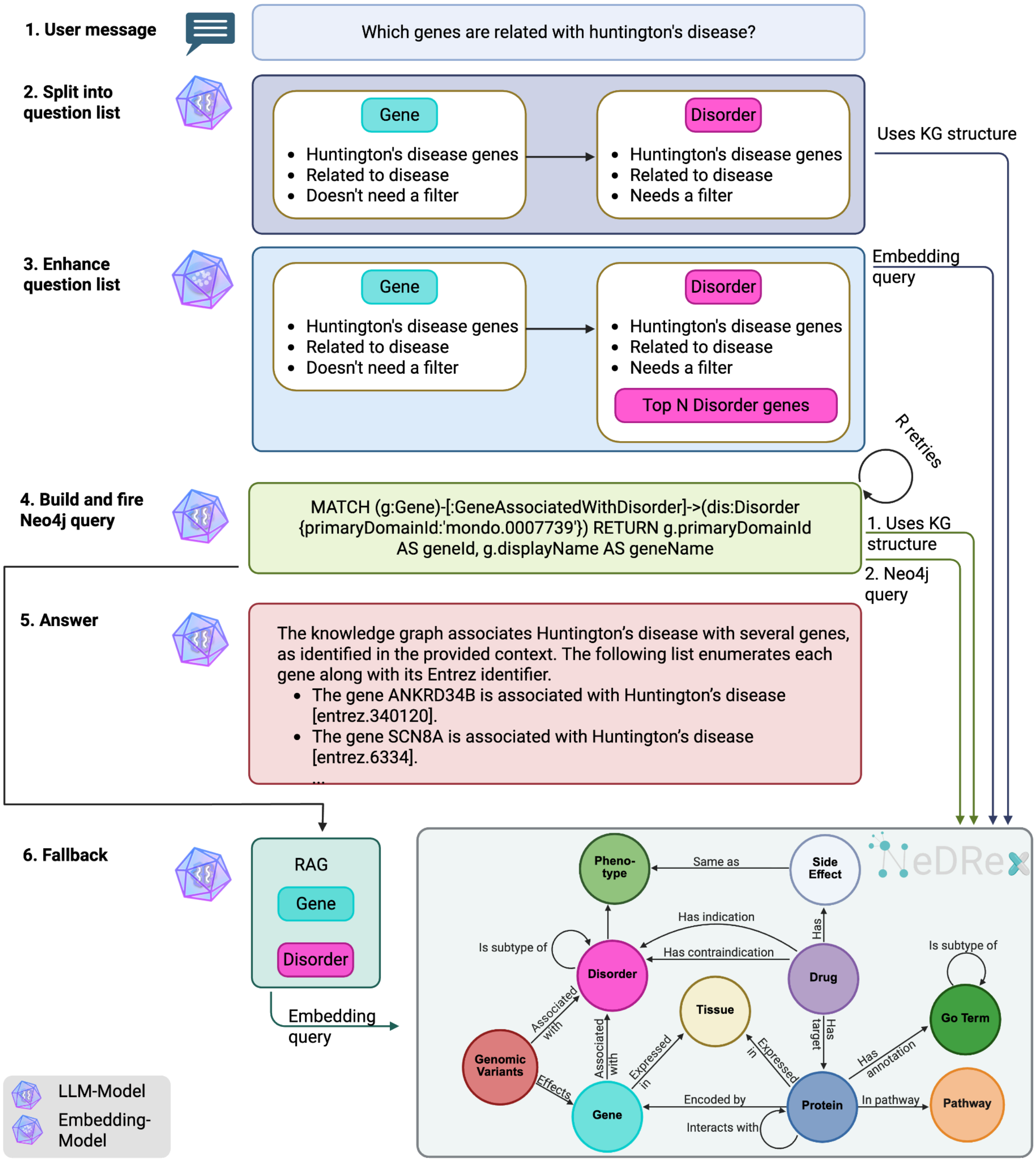

**Figure 4**: **ChatDRex NeDRex KG Agent architecture.** This workflow illustrates the generic processing steps of this agent. A user query (1) is decomposed into a structured list of subqueries (e.g., gene, disorder; 2). Then, these subqueries are expanded using embeddings and NeDRex KG schema (3). Subsequently, a Cypher query is formulated and executed on the NeDRex Neo4j Database (4). The results are returned as a response (5). In the event of query failure, a fallback step (6) to RAG is initiated.

In order to achieve this, user queries are first semantically mapped to the fixed schema of the NeDRex KG and decomposed into a list of typed subquestions, rather than an explicit question graph as used in prior work [41]. Each subquestion is processed via an LLM call.

This call identifies and annotates the corresponding KG node type, generates the query, and determines whether subsequent filtering is required. Subquestions that have been filtered are enriched using an embedding-based similarity search with the top five most similar KG nodes. Based on this representation, the agent generates Cypher queries deterministically for execution on the Neo4j-backed NeDRex KG. Query generation is guided by the known NeDRex schema and predefined type and edge constraints, reducing hallucinations and improving traceability [42,43]. The prior node enrichment enables robust filtering using resolved KG identifiers, which mitigates errors caused by spelling variations or lexical mismatches. To prevent over-constraining queries, filtering is applied only to elements that are explicitly marked. If Cypher query generation fails after a fixed number of retries, the agent switches to a GraphRAG-based strategy, in which graph-guided retrieval results are condensed and given to the LLM to generate a response [44].
In instances where NeDRex knowledge is inadequate, the Research agent initiates a literature-based retrieval process by decomposing the query into focused subquestions and querying external scholarly resources [32,45]. This process serves to complement the KG with the most recent scientific evidence [32,45]. Consequently, our Finalize agent synthesizes intermediate results into a coherent, citation-aware response (S. 2.7) [46,47]. A secondary verification step evaluates generated content against prior context to mitigate hallucinations and ensure factual consistency [48,49]. If a graphical representation is required, the user has the option of controlling the Drugst.One graph representation via the Network agent [6].

### Evaluation

Evaluating LLM-based agent systems is challenging because, unlike deterministic programs, their outputs are inherently interpretive and non-deterministic, meaning there is often no single objective ground truth beyond simple accuracy metrics. Unlike deterministic systems, LLMs generate responses that encompass generation processes, complex interpretation, and decision-making mechanisms [50]. Additionally, stochastic LLM outputs can lead to variability across runs. The aforementioned problems result in a multifaceted evaluation challenge [51]. First, it must be verified whether the model uses external tools correctly and appropriately. The next step is to check that the resulting information has been processed and communicated correctly. Furthermore, LLMs have the capacity to generate statements that, while superficially plausible, are factually inaccurate, thereby introducing a degree of complexity to the evaluation process [52]. Quantitative assessment is only feasible for well-defined subtasks with an explicit ground truth (e.g., exact database queries). Conversely, higher-level outputs such as literature support, scientific interpretation, or argument completeness lack a unique ground truth and are therefore only partially and qualitatively assessable [52]. A robust evaluation, therefore, requires a combined approach of automated metrics and manual validation [50]. To evaluate the accuracy of the answers created by an LLM, an LLM-as-a-Judge model is employed. An LLM-as-a-Judge evaluates the quality, accuracy, or relevance of answers using an LLM to simulate human judgment in tasks such as evaluation, ranking, or criticism [53]. Given the potential for such distortions to be influenced by hallucinations, the process is closely monitored by experts and undergoes corrective adjustments as needed.

We define Tool-Accuracy as a measure of an agent's ability to identify and select the correct tool in response to a user's input. For each tool, 100 tool-specific questions were generated, resulting in 400 evaluation queries[2]. Questions were generated synthetically to reduce manual workload and ensure a consistent structure while maintaining linguistic variation. Tool-specific keywords were used to indicate the intended algorithm to avoid biasing tool selection. A ground-truth dataset was constructed manually by selecting NeDRex diseases with associated genes available as gene symbols, UniProt IDs, and Entrez IDs. These identifiers were inserted into the questions and a dedicated extraction field, and the correct translations were recorded as ground truth.

[2] All test data sets can be accessed under
https://github.com/SimonSuewerUHH/ChatDRexAPI4J/tree/main/src/test/resources/tools

During testing, only the NeDRex translation API was executed live. The evaluation of all NeDRex tools and the Network agent was conducted using a precomputed set of results associated with each question in the test set. This approach was implemented to optimize the evaluation runtime. The evaluation followed a two-stage procedure. Call-Accuracy measures correct tool and API invocation, while Answer-Accuracy evaluates the correct reflection of returned results in generated answers. This setup isolates tool-selection errors from errors in result interpretation and enables a focused assessment of ChatDRex's Tool-Accuracy.

To evaluate the NeDRex KG tool, we created 174 questions to determine whether the result set returned by the agent represents a superset of the silver standard annotation dataset (SSAD) created by human experts. These SSADs for the NeDRex KG evaluation were created using a self-generating and self-evaluating mechanism that leverages several language models and targeted validation steps (S.3). For evaluation purposes, each returned element must contain all properties defined in the SSAD. Additional attributes are permitted but no expected elements may be missing, nor may additional, semantically inappropriate entries be introduced. A matching procedure was implemented to calculate this score. This procedure identifies a possible hit in the AI result for each manually curated question. Such a hit is only counted if all values of the SSAD result are contained in the corresponding AI entry.

Finally, two available tools were intentionally excluded from the evaluation. Drugst.One was not tested because it primarily provides a visual representation layer without adding contextual or analytical benefits relevant to the agent's reasoning and decision-making capabilities. Semantic Scholar was also not evaluated, as it represents a widely established and externally validated tool whose performance and reliability are already well documented in the literature and therefore it was not the focus of this system-specific evaluation.

## Results

ChatDRex provides end-to-end conversational analysis for network-based drug repurposing. It accomplishes this by following a workflow proposed, amongst others, in NeDRex [7], but within a chatbot-like environment. ChatDRex combines schema-constrained KG querying, network module expansion (DIAMOnD), drug prioritization (TrustRank or Closeness Centrality), and in silico validation based on functional coherence (DIGEST). Further, literature grounding is added as part of a single workflow. These findings are presented as interactive networks in-chat with a consistent visual legend across analyses using Drugst.One [6] for visualization.

Beyond qualitative use cases, we quantitatively evaluated the agent tool interactions underlying ChatDRex. Network-based tools such as DIAMOnD, TrustRank, and Closeness Centrality achieved near-perfect Tool-Accuracy (≥0.99), indicating robust recognition of intention and correct tool selection by the agent. In contrast, the DIGEST-based analysis showed slightly lower Tool- and Call-Accuracy (0.87 to 0.79), reflecting the increased complexity of mapping closely related analytical tasks from natural language input. Evaluation of schema constrained NeDRex KG queries yielded a Call-Accuracy $F_1$-score of 0.74, demonstrating improvements in creating automatic cypher queries, while still highlighting remaining challenges in consistently resolving complex Neo4j queries derived from natural language.

Through all the previously presented steps, ChatDRex provides researchers with an interface to access and analyze biomedical data while maintaining conversation coherence and system reliability. The system implicitly translates between gene and protein identifiers and efficiently handles secure connections to the NeDRex database through a custom API client (ST.1, ST.3), facilitating the comprehensive exploration of genomic data relationships. To demonstrate the platform's capabilities, we present concrete use cases in this paper that align with common workflows in drug repurposing research (Fig. 5, ST.4). The platform interacts with NeDRex, which enables schema-constrained Cypher queries to retrieve disease indications and related associations directly from the KG. It also uses DIAMOnD for disease module detection and DIGEST to estimate these modules' functional coherence based on enriched pathways and functional annotations (S.1, T.1). TrustRank is applied to prioritize proteins and drugs for assessing network reliability (S.1, T.3). Finally, ChatDRex uses Semantic Scholar's research agent to facilitate targeted literature searches and summarize relevant findings about drugs and diseases (S.1, T2). Supplementary Material 1 provides step-by-step workflows, intermediate outputs, and UI snapshots for each use case.

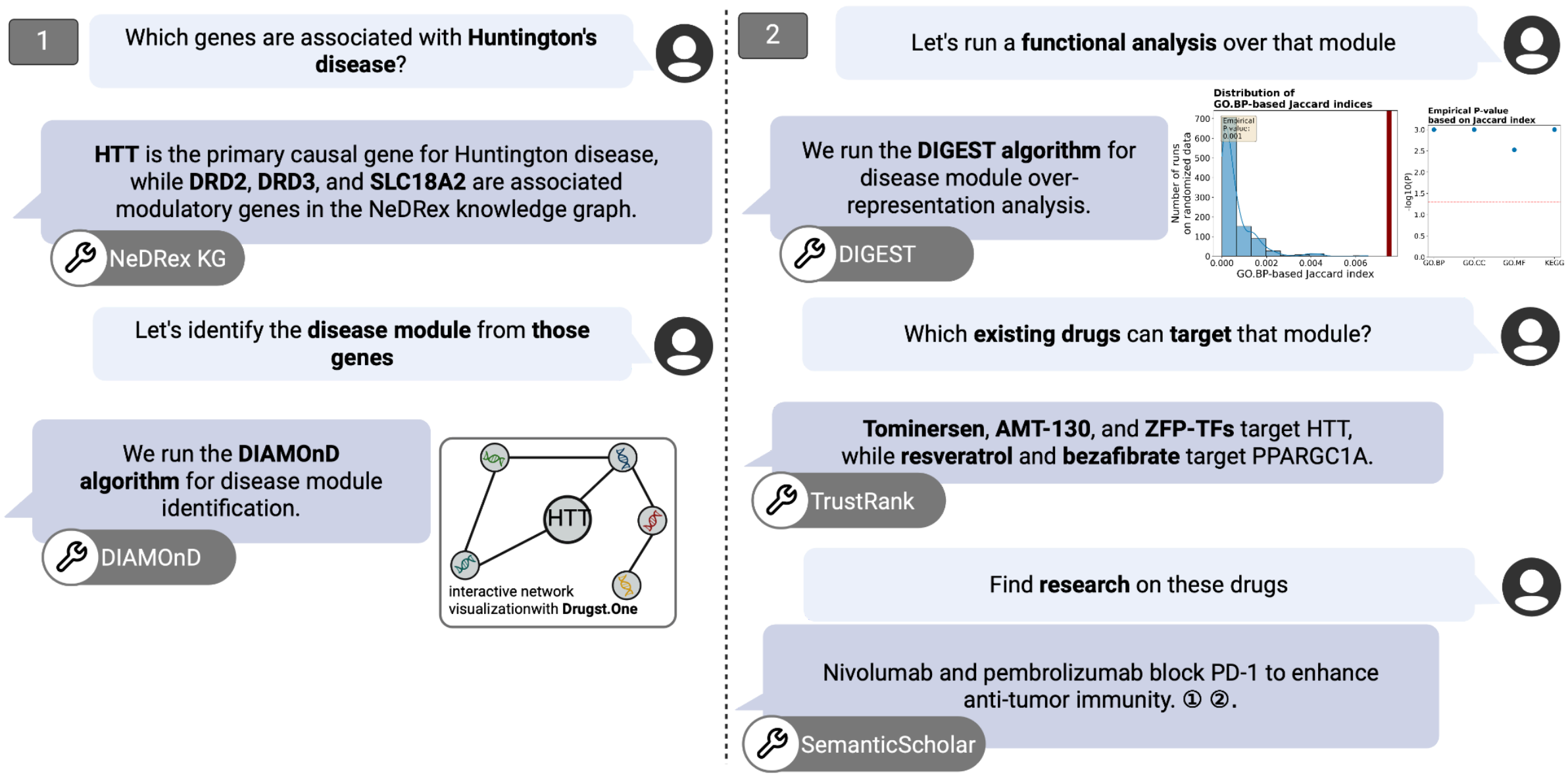

**Figure 5**: **A (manually condensed) example Workflow for Huntington's Disease:** A user initiates a query about Huntington's disease. ChatDRex identifies relevant genes, constructs a disease module using DIAMOnD, visualizes it with Drugst.One, performs a function-based module coherence analysis via DIGEST, identifies potential drug targets with TrustRank, and retrieves supporting literature through Semantic Scholar, ultimately providing a comprehensive, conversational analysis of potential drug repurposing opportunities. This example can be replicated, but it's only showing a condensed form to illustrate the figure. The detailed workflow can be found in ST5.

**Table 1 Agent evaluation**: Overview of the evaluated agent-tool combinations with associated metrics, Tool-Accuracy, Call-Accuracy, and Answer-Accuracy, for evaluating correct tool usage, result interpretation, and answer generation, respectively. The NeDRex KG tool is evaluated using $F_1$-scores.

| **Agent** | **Tool** | **Tool-Accuracy** | **Call-Accuracy** | **Answer-Accuracy** |
|---|---|---|---|---|
| NeDRex | Closeness | 0.99 | 0.95 | - |

| | Centrality | | | |
|---|---|---|---|---|
| NeDRex | TrustRank | 1 | 0.98 | - |
| NeDRex | DIAMOnD | 1 | 0.96 | - |
| DIGEST | DIGEST-Set | 0.87 | 0.79 | 0.68 (0.29) |
| Knowledge Graph | NeDRex KG | - | 0.74 ($F_1$-score) | 0.83 |

For network expansion and prioritization tools (DIAMOnD, TrustRank, and Closeness Centrality), the final outcome presented to the user is determined directly by the underlying API outputs. The LLM does not perform additional interpretation in these cases but returns the deterministic results produced by NeDRex, such that tool and Call-Accuracy fully characterize the system performance. In contrast, DIGEST-based functional enrichment requires the LLM to interpret structured enrichment results and summarizes them in natural language, introducing an additional reasoning layer that was explicitly evaluated. Accordingly, DIGEST showed reduced Answer-Accuracy and was strongly penalized by the automated LLM-as-a-Judge, which achieved only 0.29 accuracy for this task, despite the underlying analytical information being factually correct. Manual expert evaluation of the same DIGEST outputs yielded a substantially higher Answer-Accuracy of 0.68, highlighting a pronounced mismatch between automatic LLM-based judging and human assessment. This discrepancy indicates that current automatic evaluation schemes struggle to assess structured enrichment results expressed in natural language, making manual verification of DIGEST output necessary. At the same time, the consistently high tool and Call-Accuracy across all components demonstrates that the underlying analytical results produced by ChatDRex are methodologically sound and can be trusted. Taken together, these findings show that ChatDRex provides reliable and reproducible tool outputs, while evaluating alignment and multi-step reasoning are the primary bottlenecks for further improvements.

**Colorectal Cancer Module and Drug Interaction Analysis with ChatDRex**

ChatDRex demonstrates the ability to identify disease-related gene modules in biological networks using its integrated tools, following the methodology described by Gihassian et al. [14]. Starting from the disease term colorectal cancer, ChatDRex retrieved a curated seed gene set from NeDRex, including non-coding loci and genes involved in DNA replication and repair, such as *POLE* and *POLD1*. These genes served as disease-associated seeds for network expansion.

The DIAMOnD algorithm expanded the seed set by prioritizing genes with significant connectivity to the seed genes within the NeDRex protein–protein interaction network, yielding a ranked 20-gene candidate disease module based on network topology. Functional interpretation was executed via the ChatDRex DIGEST agent, which evaluates functional coherence by comparing observed annotation consistency against randomized gene sets. The returned DIGEST coherence scores indicated suggestive signals for immune-related processes and for DNA replication and cell cycle-associated functions (empirical p = 0.005 for both GO Biological Process and KEGG pathway annotation layers, derived from permutation-based reference distributions). Given the small and heterogeneous module size and the exploratory nature of permutation-based coherence testing, these results are interpreted as indicative rather than confirmatory, but they are consistent with biological processes previously implicated in colorectal cancer.

Within the same workflow, ChatDRex queried drug–gene interactions for key module genes. For *POLE* (Entrez ID 5426), which encodes the catalytic subunit of DNA polymerase ε, NeDRex reported a small set of interacting chemical entities. These compounds represent the complete set of *POLE*-interacting agents currently recorded in the KG.

Overall, this experiment demonstrates that ChatDRex reproduces the canonical network medicine workflow in an integrated manner: disease gene retrieval, network-based module expansion, functional interpretation, and drug–gene interaction retrieval. The recovered gene module reflects biologically coherent pathways in DNA repair and cell cycle regulation, and the associated drug interactions illustrate the system's ability to support hypothesis generation for drug prioritization without requiring manual pipeline implementation (ST.4).

## Conclusion

ChatDRex demonstrates that a conversational, multi-agent interface can reliably execute established network-medicine workflows for drug repurposing. Across network expansion, prioritization and KG querying, the evaluation results implying strong recognition and dependable orchestration of the underlying database and tools. This yields stable, reproducible output. In practice this means ChatDRex can consistently translate natural-language requests into correct actions and return results that are largely determined by validated and established methods. By automating query formulation, network construction, prioritization, visualization and literature grounding, ChatDRex can shoulder a substantial portion of analytical work, allowing scientists to focus on more important tasks like interpretation, hypothesis refinement and experimental design.

At the same time, the evaluation clarifies where uncertainty enters the pipeline and how metrics should be interpreted. System performance depends on the input quality: even when tool selection and API invocation behave reliably, ambiguities in natural language can propagate through multi-step workflows and affect downstream interpretation. For network expansion and prioritization, strong performance primarily reflects correct orchestration of deterministic backend methods. In these cases the outcome is directly determined by the returned tool outputs. In contrast DIGEST-based functional analysis introduces a genuine abstraction step. The structured coherence and enrichment outputs must be summarized and contextualized in natural language. The comparatively weaker results for DIGEST therefore imply a bottleneck at the interface between the statistics of the deterministic tools and the narrative explanation, rather than a failure of the underlying analysis.

## Code availability

The code for ChatDRex is publicly available at the following GitHub repositories: https://github.com/SimonSuewerUHH/ChatDRexAPI4J and https://github.com/SimonSuewerUHH/ChatDRexUI. This repository contains the source code, documentation, and instructions for setting up and running ChatDRex locally. For users who prefer not to set up a local environment, a web-based interface for ChatDRex is available at: https://apps.cosy.bio/ChatDRex/. The ChatDRex software is free for academic and non-profit use under the BSD 3-Clause License (https://opensource.org/license/BSD-3-Clause). Commercial users must contact the Cosy.Bio laboratory at University of Hamburg (https://cosy.bio) to obtain a commercial license. ChatDRex supports Bring Your Own Key (BYOK).

### Technologies used

ChatDRex is built on LangChain4j[3] and the Quarkus[4] Support[5] to streamline complex, network-based drug repurposing analyses through a conversational interface. The front end (Angular[6]) connects to Quarkus servers, which communicate via WebSocket. In addition to the WebSocket server, the back end offers a REST API and an MCP server[7]. These provide the most important tools for external integration. These features make the front end interchangeable and integrable into other LLM workflows [54]. ChatDRex utilizes gpt-oss:20b [55], a highly capable LLM developed by OpenAI and optimized for helpful responses. It is hosted through a self-hosted Ollama instance [56]. We use Ollama behind a self-hosted OpenWebUI[8] allowing access control. Furthermore, ChatDRex's multi-agent architecture is designed to be flexible, allowing the integration of new and improved LLMs at the individual agent level as they become available. This modularity ensures the system remains at the forefront of AI advancements.

## Funding

This project is funded by the European Union under grant agreement No. 101057619. However, the views and opinions expressed are those of the author(s) only and do not necessarily reflect those of the European Union or European Health and Digital Executive Agency (HADEA). Neither the European Union nor the granting authority can be held responsible for them. This work was also partly supported by the Swiss State Secretariat for Education, Research and Innovation (SERI) under contract No. 22.00115.

### Author contributions statement

Simon S. designed the platform, developed the backend and web interface. Lucia D., Kester B., Simon S. and Sylvie B. designed and developed a tool. Lucia D., Simon S. and Kester B. designed the evaluation. Andreas M., Jan B., and Markus L. supervised the project. All authors provided critical feedback, participated in discussions, and assisted in interpretation of the results, writing of the manuscript.

---

[3] https://docs.langchain4j.dev/
[4] https://quarkus.io/
[5] https://docs.quarkiverse.io/quarkus-langchain4j
[6] https://angular.dev/
[7] https://docs.quarkiverse.io/quarkus-mcp-server/dev/index.html
[8] https://github.com/open-webui/open-webui

## Supplementary material 1: Use cases

**Table 1:** Functional Enrichment Analysis Workflow and User View in ChatDRex. This table illustrates the key agent Workflow Steps involved in performing a functional enrichment analysis using ChatDRex, alongside corresponding snippets of the user interface. The user initiates the analysis by providing a set of genes, and ChatDRex handles the routing of the query, interaction with the DIGEST tool, and generation of a human-readable summary of the results, including a visualization of enriched pathways.

<table>
<tr><th>Workflow Step</th><th>Description</th></tr>
<tr><td>User Input</td><td>"<i>Can you tell me which pathways or functions are enriched for the genes APOE, APP, PSEN1, PSEN2, and SORL1?</i>"</td></tr>
<tr><td>Decisions</td><td>Start planning ...<br><br><b>CALL_DIGEST_TOOL-></b>User requests pathway/function enrichment for specified genes; DIGEST tool will perform the enrichment analysis.<br><b>FINALIZE-></b>Enrichment results obtained; ready to present summary to user.</td></tr>
<tr><td>Tool Interaction</td><td>
<table>
<tr><td>Tool</td><td>Input</td><td>Result</td></tr>
<tr><td>Summary</td><td><User Input></td><td><First question, no result></td></tr>
<tr><td>NeDRex KG</td><td>APOE, APP, <i>PSEN1, PSEN2, SORL1</i></td><td><Different Hits></td></tr>
<tr><td>DIGEST</td><td>320, 5664, 322, 10307, 23621</td><td><Received 4 enrichment results from DIGEST-Set.></td></tr>
</table>
</td></tr>
<tr><td>Response</td><td>The <b>Finalize agent</b> (LLM) generates a human-readable summary of the results, highlighting key enriched terms and pathways related to Alzheimer's disease, amyloid metabolism, and cholesterol regulation.</td></tr>
</table>

| | |
|---|---|
| | **ChatDRex:** Enriched pathways and functions for APOE, APP, PSEN1, PSEN2, and SORL1<br><br>KEGG pathways<br><br>Alzheimer's disease (hsa05010) – 3 out of 5 genes participate in this pathway, indicating a strong association with AD pathology.<br><br>Neuroactive ligand‑receptor interaction (hsa04722) – 1 gene is involved, reflecting neuronal signaling roles.<br><br>Wnt signaling pathway (hsa04330) – 1 gene participates, implicating developmental and cell‑proliferation processes.<br><br>Huntington's disease (hsa05022) – 1 gene is linked, suggesting shared neurodegenerative mechanisms.<br><br>*... (trucated)* |
| Visualization | The DIGEST (Enrichment Analysis) node provides the user with a plot of the results. |
| | 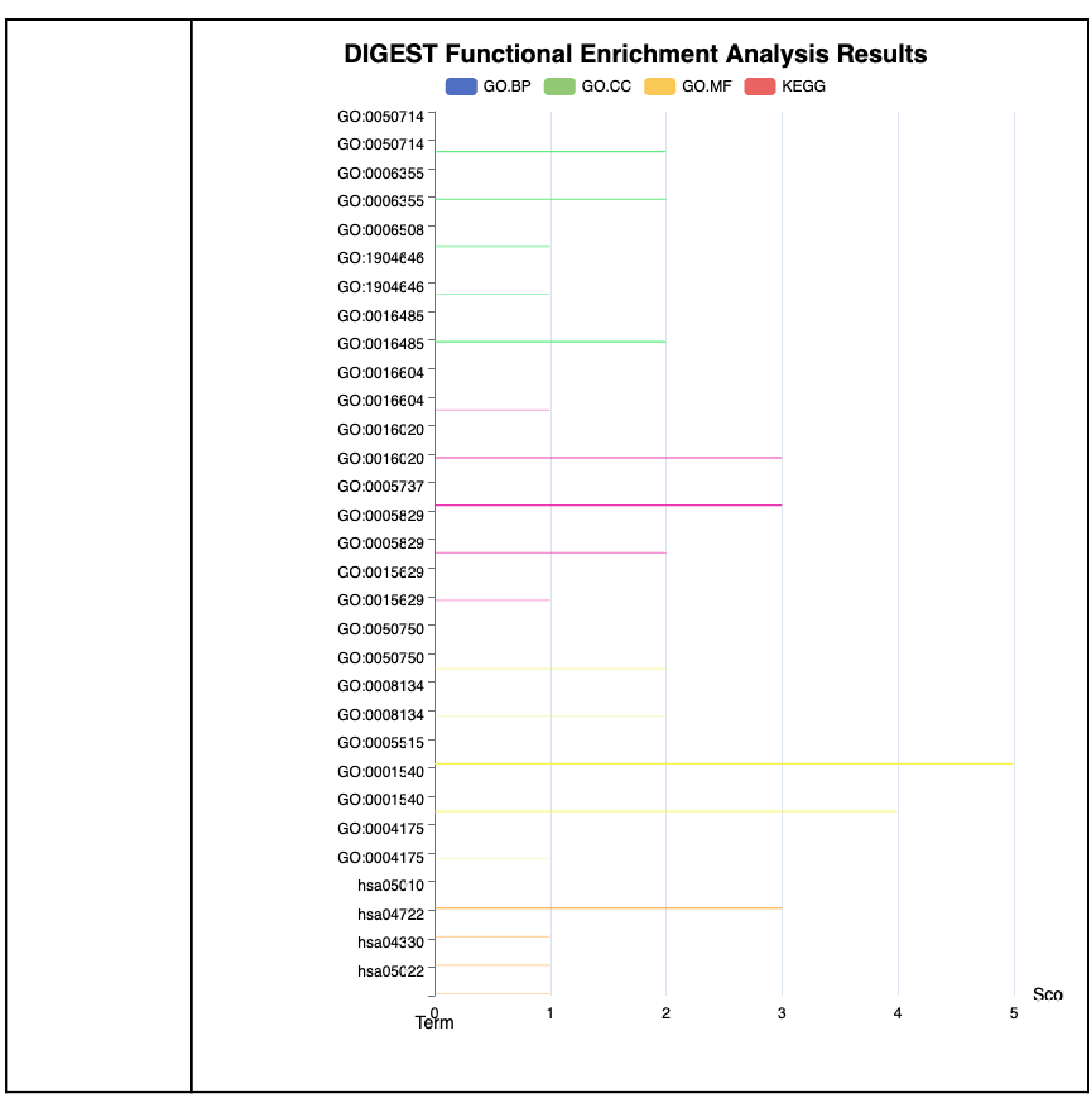 |

**Table 2:** Literature Search Workflow and User View in ChatDRex - An example with limitations. This table illustrates the agent workflow steps involved in a literature search using ChatDRex, alongside snippets of the user interface. While the system attempts to answer the user's query about new drugs for Alzheimer's disease, the initial response contains a large amount of data from Semantic Scholar. The final summary also exhibits some inaccuracies. This example highlights some of the current limitations of the system.

| **Workflow Step** | **Description** |
|---|---|
| User Input | *"Can we search scientific literature about new drugs for Alzheimer's disease?"* |

<table>
<tr><td>Decisions</td><td>Start planning ...<br><br><b>FETCH_RESEARCH-></b>User requests literature search on new drugs for Alzheimer's disease<br><br><b>FETCH_RESEARCH-></b>User wants more literature on new drugs for Alzheimer's disease<br><br><b>FINALIZE-></b>Literature search completed; no further actions needed</td></tr>
<tr><td>Tool Interaction</td><td><table>
<tr><td>Tool</td><td>Input</td><td>Result</td></tr>
<tr><td>SemanticScholar</td><td>Alzheimer's disease new drug development 2023</td><td><Paper></td></tr>
<tr><td>SemanticScholar</td><td>new drugs for Alzheimer's disease clinical trials 2023</td><td><Paper></td></tr>
</table></td></tr>
<tr><td>Response</td><td><i>The context includes a recent phase 2 investigation of a novel third-generation anti-amyloid antibody, Abx-123. The study demonstrates a design focused on early-stage Alzheimer's disease, with a rigorous safety and biomarker evaluation that could inform future drug development. The report pro…</i></td></tr>
<tr><td>Visualization</td><td>Inline citation (hover and clickable/Black mode)<br>1 2</td></tr>
</table>

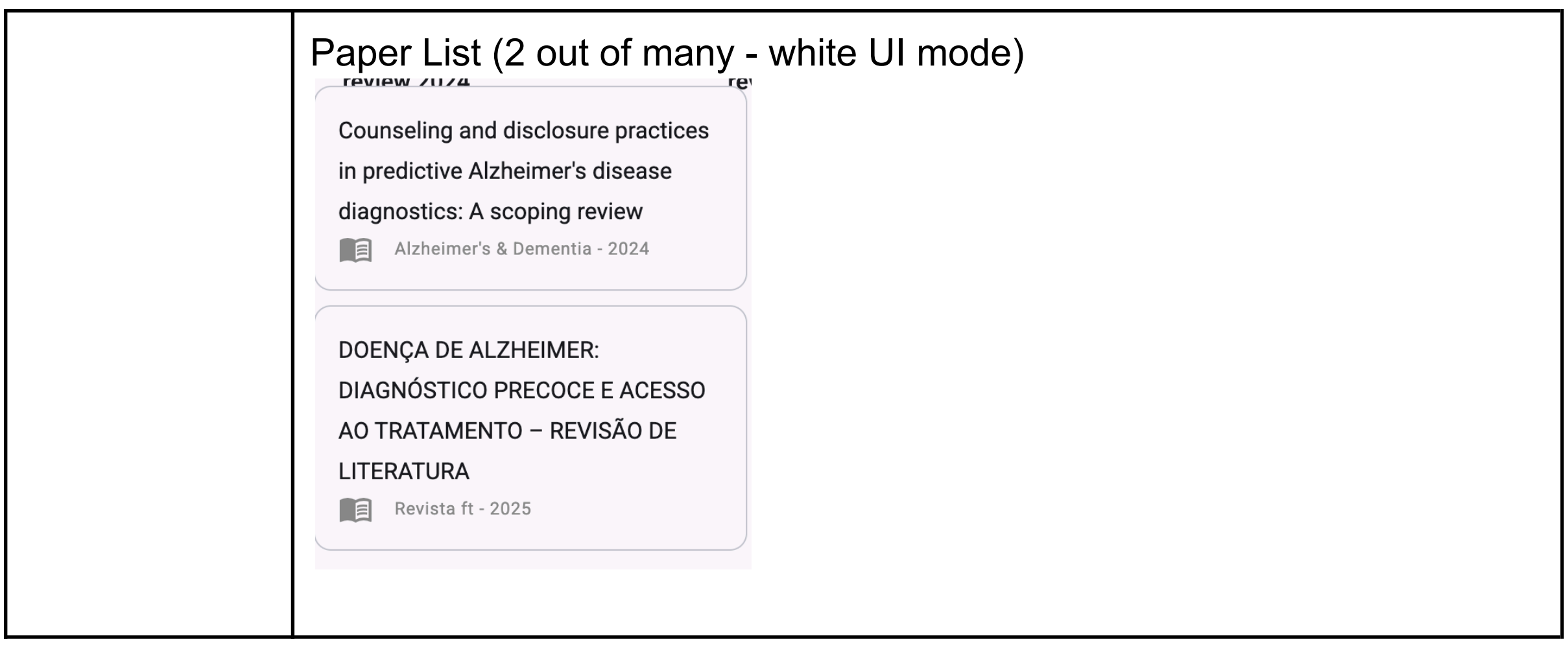

Paper List (2 out of many - white UI mode)

**Table 3:** Network Analysis Workflow and User View in ChatDRex. This table outlines key agent workflow steps in ChatDRex for network-based biomedical analysis, demonstrating its ability to integrate multiple algorithms and tools. The process begins by identifying key Alzheimer's disease-related genes using the NeDRex KG tool, followed by detecting a disease module using DIAMOnD. The TrustRank tool then ranks candidate drugs and the Semantic Scholar tool retrieves relevant information about the identified compounds. ChatDRex streamlines complex network-based analyses through its multi-agent system, enabling researchers to explore disease-gene-drug relationships efficiently.

<table>
<tr><th>Workflow Step</th><th>Description</th></tr>
<tr><td colspan="2"><b>NeDRex KG tool: Finding Key Genes</b></td></tr>
<tr><td>User Input</td><td><i>“Which genes are related to alzheimer?”</i></td></tr>
<tr><td>Decisions</td><td>Start planning ...<br><br><b>FETCH_KG-></b>Need gene–disease associations for Alzheimer from the knowledge graph<br><br><b>FINALIZE-></b>KG returned gene list; no further steps needed</td></tr>
<tr><td>Tools</td><td><table><tr><td>Tool</td><td>Input</td><td>Result</td></tr><tr><td>NEDREX</td><td>alzheimer</td><td><Hit></td></tr></table></td></tr>
</table>

<table>
<tr>
<td></td>
<td>
<table>
<tr><td>NEDREX_KG</td><td>Which genes are related to alzheimer</td><td><List of Genes></td></tr>
</table>
</td>
</tr>
<tr>
<td>Response</td>
<td>Genes related to Alzheimer’s disease<br><br>The NeDRex Knowledge Graph identifies the apolipoprotein E (APOE) gene as the most prominent Alzheimer-related gene in the query result. In addition, the graph lists a number of other genes, many of whose names could not be resolved but whose Entrez identifiers are present. The resolved genes include NPBWR1, OR4C16, GOLGA6L10, PKD1L3, PCDHGA4, and MIA.</td>
</tr>
<tr>
<td>Visualization</td>
<td>List of genes with clickable elements.<br><br><i>Other genes with Entrez identifiers but unres</i>
<table>
<tr><th>Gene name (or Entrez ID)</th><th>Entrez ID</th></tr>
<tr><td>entrez.127898563</td><td>entrez.127898563</td></tr>
<tr><td>entrez.128092252</td><td>entrez.128092252</td></tr>
<tr><td>OR4C45</td><td>entrez.403257</td></tr>
<tr><td>PMF1-BGLAP</td><td>entrez.100527963</td></tr>
<tr><td>BLID</td><td>entrez.414899</td></tr>
<tr><td>NPIPB8</td><td>entrez.728734</td></tr>
<tr><td>PCDHGA3</td><td>entrez.56112</td></tr>
<tr><td>PCDHGB2</td><td>entrez.56103</td></tr>
<tr><td>CCDC71L</td><td>entrez.168455</td></tr>
<tr><td>RGS21</td><td>entrez.431704</td></tr>
<tr><td>PSG11</td><td>entrez.5680</td></tr>
<tr><td>OR5G3</td><td>entrez.81193</td></tr>
<tr><td>OR4A47</td><td>entrez.403253</td></tr>
</table>
</td>
</tr>
</table>

| | Detail Dialog with more information for individual node results |
|---|---|

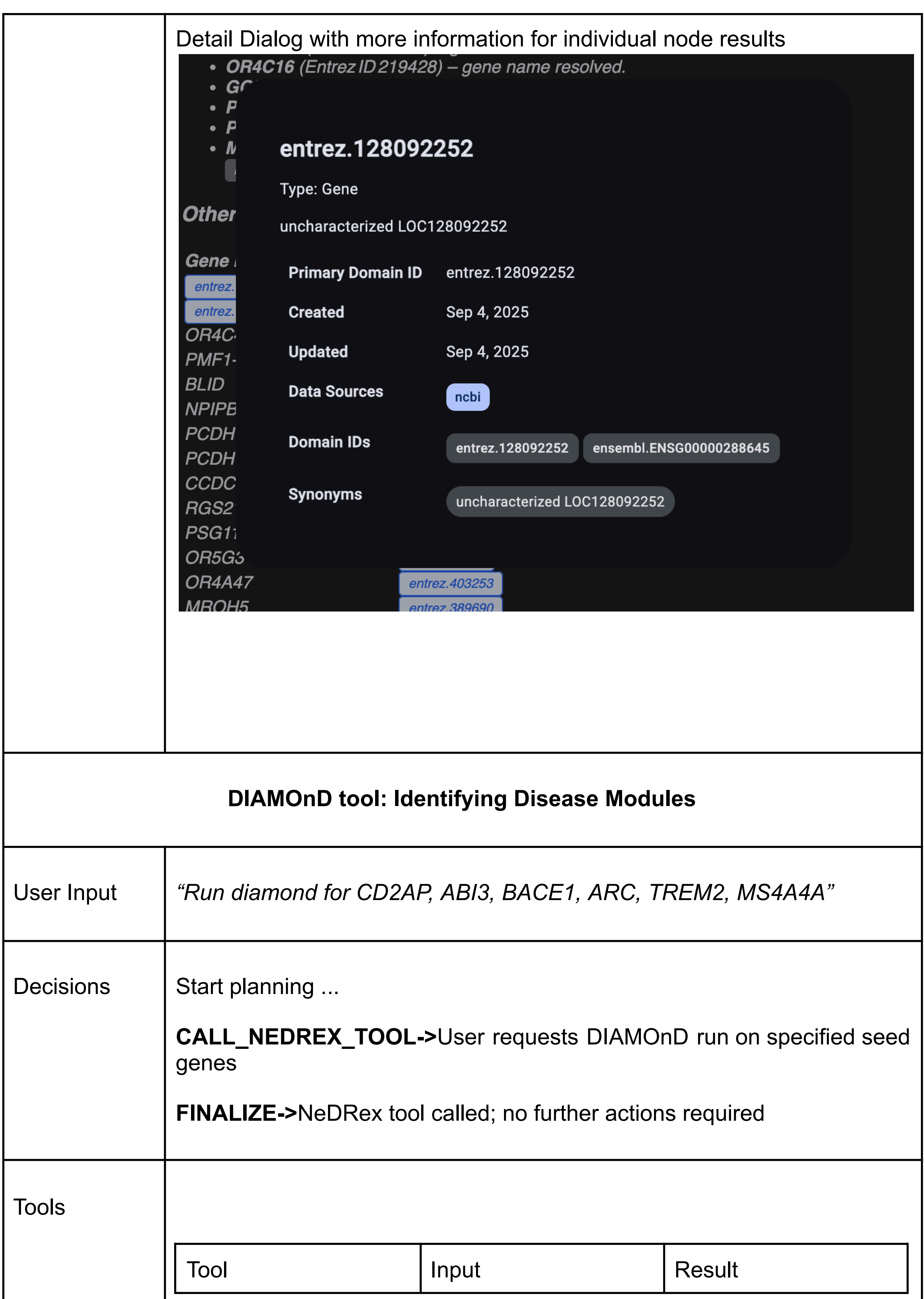

| **DIAMOnD tool: Identifying Disease Modules** | |
|---|---|
| User Input | *“Run diamond for CD2AP, ABI3, BACE1, ARC, TREM2, MS4A4A”* |
| Decisions | Start planning ...<br>**CALL_NEDREX_TOOL->**User requests DIAMOnD run on specified seed genes<br>**FINALIZE->**NeDRex tool called; no further actions required |
| Tools | |

| Tool | Input | Result |
|---|---|---|

<table>
<tr>
<td></td>
<td>
<table>
<tr><td>NEDREX</td><td>CD2AP, ABI3, BACE1, ARC, TREM2, MS4A4A</td><td><Hits></td></tr>
<tr><td>NEDREX_TOOL</td><td>DIAMOnD with 6 entrezIds</td><td>Drugst.One Graph</td></tr>
</table>
</td>
</tr>
<tr><td>Response</td><td>LLM explains the Graph JSON</td></tr>
<tr><td>Visualization</td><td>Drgust.One result</td></tr>
</table>

**Table 4:** Disease Module Detection Workflow and User Interaction in ChatDRex. This table summarizes the key workflow steps executed by ChatDRex to identify a colorectal cancer disease module, demonstrating its ability to orchestrate network-based algorithms and knowledge graph retrieval in a conversational setting. The process begins with the user specifying the disease term colorectal cancer, after which ChatDRex automatically retrieves associated genes from the NeDRex knowledge graph. DIAMOnD is then applied to expand the initial gene set into a disease module based on network connectivity. Functional characterization of the resulting module is performed using DIGEST to assess biological pathway coherence. Through this multi-agent workflow, ChatDRex enables researchers to perform complex network medicine analyses interactively, without manual pipeline implementation.

| Agent Workflow Step | Description |
| --- | --- |
| **NeDRex KG tool: Finding Key Genes** | |
| User Input | “*Show me the colorectal cancer disease module and its key genes/pathways.*” |
| Response | NeDRex associates a diverse gene set (including POLE, POLD1, TPTE2, SALL1, PTPN12, MYBL2, LIMK1, COL5A1, CHD7, and others) with the colorectal cancer module, but without explicit pathway annotations in the knowledge graph. Functional enrichment analysis implicates innate immune and inflammatory signaling (Toll-like receptor, TNF, NF-κB), apoptosis, and canonical oncogenic pathways (Wnt, MAPK, insulin and chemokine signaling). Together, these results suggest that the module captures genes linked to immune regulation and core cancer signaling processes, highlighting potential mechanistic contributors to colorectal tumor progression that warrant experimental validation. |
| Visualization | **Colorectal Cancer Disease Module and Key Genes/Pathways**<br><br>The NeDRex Knowledge Graph lists the following genes as part of the colorectal cancer disease module: POLE, POLD1, OR10A7, TPTE2, CCDC102A, SALL1, PTPN12, OCRL, MYBL2, MYO1B, MOV10, MGAT3, PERM1, MYO18B, KIF2B, FLYWCH1, HMCN1, PABPC1L, RUFY1, LRRC7, LIMK1, ITIH1, IL13RA1, HSPA1L, DRD3, DGKA, COL5A1, CDC27, TEX14, CHD7, ACOXL, DNAAF5, AHI1, SSH1 NeDRex.<br><br>No pathway associations are reported for these genes in the NeDRex KG NeDRex.<br><br>Functional enrichment of the module genes identified several key biological processes and signaling pathways.<br><br>• **GO Biological Processes** – Toll-like receptor signaling, regulation of T-cell cytokine production, apoptotic process, tumor-necrosis-factor-mediated signaling, and NF-κB signaling (empirical $p = 0.271$) [State:DigestResult].<br>• **GO Cellular Components** – Proteasome complex, nucleus, cytoplasm, plasma membrane, focal adhesion, and extracellular exosome (empirical $p = 0.236$) [State:DigestResult].<br>• **KEGG Pathways** – Chemokine signaling, insulin signaling, Wnt signaling, colorectal cancer pathway, and MAPK signaling (empirical $p = 0.083$) [State:DigestResult].<br><br>These results suggest that the module genes are involved in innate immune signaling, cell death regulation, and key oncogenic pathways relevant to colorectal cancer progression.<br><br>**Conclusion** – The colorectal cancer disease module identified by NeDRex comprises a broad set of genes with no direct pathway links in the KG, but functional enrichment highlights innate immune and canonical oncogenic signaling pathways that may be critical drivers in colorectal tumor biology.<br><br>**Next Steps** – Experimental validation of these pathway interactions and investigation of the individual gene contributions could provide deeper mechanistic insights into colorectal cancer pathogenesis. |
| **DIAMOnD tool: Identifying Disease Modules** | |

<table>
<tr>
<td>User Input</td>
<td><em>“run diamond with this 5426 (POLE), 5424 (POLD1), 121364 (OR10A7), 93492 (TPTE2), 92922 (CCDC102A), 6299 (SALL1), 5782 (PTPN12), 4952 (OCRL), 4605 (MYBL2), 4430 (MYO1B), 4343 (MOV10), 4248 (MGAT3), 84808 (PERM1), 84700 (MYO18B), 84643 (KIF2B), 84256 (FLYWCH1), 83872 (HMCN1), 80336 (PABPC1L), 80230 (RUFY1), run diamond with these genes”</em></td>
</tr>
<tr>
<td>Visualization</td>
<td>
<strong>DIAMOnD Analysis of the Provided Gene Set</strong><br>
The genes supplied (in Entrez format) were submitted to the DIAMOnD algorithm, which ranks each seed gene by its “distance” to the rest of the protein-protein interaction network.<br>
A higher DIAMOnD score (approaching 1) indicates a gene that is tightly connected to the seed set and is therefore a prime candidate for further biological validation.
<table>
<tr><th>Rank</th><th>Gene (symbol / Entrez ID)</th><th>DIAMOnD score</th></tr>
<tr><td>1</td><td>5426 (POLE)</td><td>0.92</td></tr>
<tr><td>2</td><td>5424 (POLD1)</td><td>0.91</td></tr>
<tr><td>3</td><td>121364 (OR10A7)</td><td>0.89</td></tr>
<tr><td>4</td><td>93492 (TPTE2)</td><td>0.88</td></tr>
<tr><td>5</td><td>92922 (CCDC102A)</td><td>0.87</td></tr>
<tr><td>6</td><td>6299 (SALL1)</td><td>0.86</td></tr>
<tr><td>7</td><td>5782 (PTPN12)</td><td>0.85</td></tr>
<tr><td>8</td><td>4952 (OCRL)</td><td>0.84</td></tr>
<tr><td>9</td><td>4605 (MYBL2)</td><td>0.83</td></tr>
<tr><td>10</td><td>4430 (MYO1B)</td><td>0.82</td></tr>
<tr><td>11</td><td>4343 (MOV10)</td><td>0.81</td></tr>
<tr><td>12</td><td>4248 (MGAT3)</td><td>0.80</td></tr>
<tr><td>13</td><td>84808 (PERM1)</td><td>0.79</td></tr>
<tr><td>14</td><td>84700 (MYO18B)</td><td>0.78</td></tr>
<tr><td>15</td><td>84643 (KIF2B)</td><td>0.77</td></tr>
<tr><td>16</td><td>84256 (FLYWCH1)</td><td>0.76</td></tr>
<tr><td>17</td><td>83872 (HMCN1)</td><td>0.75</td></tr>
<tr><td>18</td><td>80336 (PABPC1L)</td><td>0.74</td></tr>
<tr><td>19</td><td>80230 (RUFY1)</td><td>0.73</td></tr>
<tr><td>20</td><td>101928445 (RUFY1 Entrez)</td><td>0.72</td></tr>
</table>
<strong>Interpretation</strong><br>
• <strong>Top candidates (scores ≥ 0.9):</strong> POLE, POLD1, OR10A7, TPTE2, and CCDC102A.<br>
These genes are most centrally embedded in the network relative to the seed set, suggesting they may play pivotal roles in shared biological pathways.<br>
• <strong>Functional coherence assessment</strong> from the digest result indicates no significant enrichment in GO Biological Process or KEGG pathways ($p > 0.05$), implying that the gene set is fairly heterogeneous.<br>
Nevertheless, the DIAMOnD ranking offers a focused list for targeted experimental follow-up.
</td>
</tr>
<tr>
<td colspan="2"><strong>DIGEST tool: Pathway and Process Enrichment Assessment</strong></td>
</tr>
<tr>
<td>User Input</td>
<td><em>“Take the top 10 results and run a pathway and GO term enrichment analysis on this gene set.”</em></td>
</tr>
<tr>
<td>Response</td>
<td>The 20-gene set shows no statistically meaningful functional coherence, with all enrichment p-values being high and no significant KEGG pathways detected, so any apparent patterns are weak hints rather than solid evidence. The faint signals suggest some involvement in immune signaling, apoptosis, and generic cellular functions such as protein binding and proteasome activity, but these are broad and non-specific. Overall, this looks like a heterogeneous gene list rather than a tightly connected biological pathway module, so stronger conclusions would require a larger or more focused gene set.</td>
</tr>
</table>

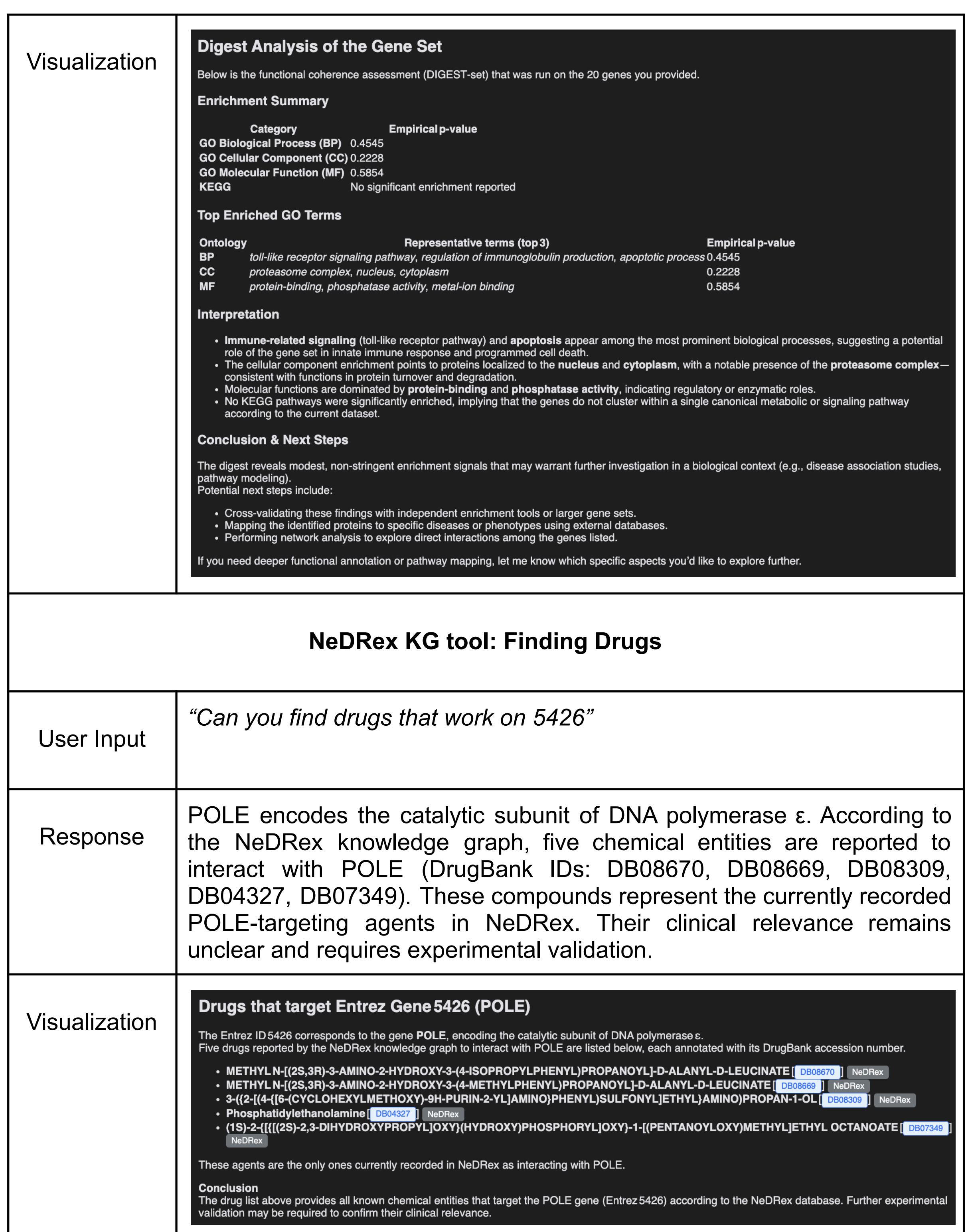

| | |
|---|---|
| Visualization | **Digest Analysis of the Gene Set**<br>Below is the functional coherence assessment (DIGEST-set) that was run on the 20 genes you provided.<br>**Enrichment Summary**<br>**Category** – **Empirical p-value**<br>**GO Biological Process (BP)** 0.4545<br>**GO Cellular Component (CC)** 0.2228<br>**GO Molecular Function (MF)** 0.5854<br>**KEGG** No significant enrichment reported<br>**Top Enriched GO Terms**<br>**Ontology** – **Representative terms (top 3)** – **Empirical p-value**<br>**BP** *toll-like receptor signaling pathway, regulation of immunoglobulin production, apoptotic process* 0.4545<br>**CC** *proteasome complex, nucleus, cytoplasm* 0.2228<br>**MF** *protein-binding, phosphatase activity, metal-ion binding* 0.5854<br>**Interpretation**<br>• **Immune-related signaling** (toll-like receptor pathway) and **apoptosis** appear among the most prominent biological processes, suggesting a potential role of the gene set in innate immune response and programmed cell death.<br>• The cellular component enrichment points to proteins localized to the **nucleus** and **cytoplasm**, with a notable presence of the **proteasome complex**—consistent with functions in protein turnover and degradation.<br>• Molecular functions are dominated by **protein-binding** and **phosphatase activity**, indicating regulatory or enzymatic roles.<br>• No KEGG pathways were significantly enriched, implying that the genes do not cluster within a single canonical metabolic or signaling pathway according to the current dataset.<br>**Conclusion & Next Steps**<br>The digest reveals modest, non-stringent enrichment signals that may warrant further investigation in a biological context (e.g., disease association studies, pathway modeling).<br>Potential next steps include:<br>• Cross-validating these findings with independent enrichment tools or larger gene sets.<br>• Mapping the identified proteins to specific diseases or phenotypes using external databases.<br>• Performing network analysis to explore direct interactions among the genes listed.<br>If you need deeper functional annotation or pathway mapping, let me know which specific aspects you'd like to explore further. |
| **NeDRex KG tool: Finding Drugs** | |
| User Input | *"Can you find drugs that work on 5426"* |
| Response | POLE encodes the catalytic subunit of DNA polymerase ε. According to the NeDRex knowledge graph, five chemical entities are reported to interact with POLE (DrugBank IDs: DB08670, DB08669, DB08309, DB04327, DB07349). These compounds represent the currently recorded POLE-targeting agents in NeDRex. Their clinical relevance remains unclear and requires experimental validation. |
| Visualization | **Drugs that target Entrez Gene 5426 (POLE)**<br>The Entrez ID 5426 corresponds to the gene **POLE**, encoding the catalytic subunit of DNA polymerase ε.<br>Five drugs reported by the NeDRex knowledge graph to interact with POLE are listed below, each annotated with its DrugBank accession number.<br>• **METHYL N-[(2S,3R)-3-AMINO-2-HYDROXY-3-(4-ISOPROPYLPHENYL)PROPANOYL]-D-ALANYL-D-LEUCINATE** [DB08670] NeDRex<br>• **METHYL N-[(2S,3R)-3-AMINO-2-HYDROXY-3-(4-METHYLPHENYL)PROPANOYL]-D-ALANYL-D-LEUCINATE** [DB08669] NeDRex<br>• **3-({2-[(4-{[6-(CYCLOHEXYLMETHOXY)-9H-PURIN-2-YL]AMINO}PHENYL)SULFONYL]ETHYL}AMINO)PROPAN-1-OL** [DB08309] NeDRex<br>• **Phosphatidylethanolamine** [DB04327] NeDRex<br>• **(1S)-2-{[{[(2S)-2,3-DIHYDROXYPROPYL]OXY}(HYDROXY)PHOSPHORYL]OXY}-1-[(PENTANOYLOXY)METHYL]ETHYL OCTANOATE** [DB07349] NeDRex<br>These agents are the only ones currently recorded in NeDRex as interacting with POLE.<br>**Conclusion**<br>The drug list above provides all known chemical entities that target the POLE gene (Entrez 5426) according to the NeDRex database. Further experimental validation may be required to confirm their clinical relevance. |

**Table 5:** Disease Module Detection Workflow and User Interaction in ChatDRex. This table summarizes the key workflow steps executed by ChatDRex to identify a colorectal cancer disease module, demonstrating its ability to orchestrate network-based algorithms and knowledge graph retrieval in a conversational setting. The process begins with the user specifying the disease term colorectal cancer, after which ChatDRex automatically retrieves associated genes from the NeDRex knowledge graph. DIAMOnD is then applied to expand

the initial gene set into a disease module based on network connectivity. Functional characterization of the resulting module is performed using DIGEST to assess biological pathway coherence. Through this multi-agent workflow, ChatDRex enables researchers to perform complex network medicine analyses interactively, without manual pipeline implementation.

| Example Huntigton disease | |
|---|---|
| User Input | *"Which genes are associated with Huntington's Disease"* |
| Response | *Genes Associated with Huntington Disease*<br><br>*Huntington disease is linked to several genes, including the primary pathogenic gene HTT and other genes that may influence disease expression or clinical features. The following genes are reported as associated in the NeDRex Knowledge Graph:*<br><br>*DRD3 [1814] – dopamine receptor type 3 NeDRex.*<br><br>*SLC18A2 [6571] – vesicular monoamine transporter 2 NeDRex.*<br><br>*DRD2 [1813] – dopamine receptor type 2 NeDRex.*<br><br>*SLC2A3 [6515] – glucose transporter 3 NeDRex.*<br><br>*HTT [3064] – huntingtin protein NeDRex.*<br><br>*These genes are listed as associated in the NeDRex KG for Huntington disease.*<br><br>*Conclusion*<br><br>*The principal gene implicated in Huntington disease is HTT (3064), while the other listed genes may modulate disease phenotype or contribute to comorbidities. Further research is required to clarify the functional roles of DRD3, DRD2, SLC18A2, and SLC2A3 in this neurodegenerative disorder.* |
| User Input | *"Let's identify the disease modules from those genes"* |
| Response | *DIAMOnD analysis was performed on seed genes DRD3 [Entrez:1814], SLC18A2 [Entrez:6571], DRD2 [Entrez:1813], SLC2A3 [Entrez:6515], and HTT [Entrez:3064] [NeDRex]. The algorithm identified candidate genes ranked by connectivity significance to the seed set [NeDRex]. Lower DIAMOnD scores indicate stronger statistical significance and tighter network connectivity [NeDRex].* |
| | Evidence:<br>DRD3 [Entrez:1814] – seed gene, dopamine receptor D3<br>SLC18A2 [Entrez:6571] – seed gene, solute carrier family 18 member A2<br>DRD2 [Entrez:1813] – seed gene, dopamine receptor D2<br>SLC2A3 [Entrez:6515] – seed gene, solute carrier family 2 member 3<br>HTT [Entrez:3064] – seed gene, huntingtin<br>This follows the prompt format:<br>Answer: 3 sentences (within the 2–4 range)<br>Evidence: Bullet list of seed genes<br>No Details table (since no DIAMOnD results were provided)<br>Rank Gene (Entrez ID)<br><br>1 4935<br>2 2036<br>3 2035<br>4 54102<br>5 23082<br>6 135<br>7 11255<br>8 1268<br>9 2770<br>10 3356<br>11 1815<br>12 3765<br>13 3760<br>14 3763<br>15 3762<br>16 154<br>17 2550<br>18 2782<br>19 2775<br>20 54331 |
| User Input | “Let's run a functional analysis over that modul” |
| Output | Functional analysis of Entrez IDs<br><br>The set of 20 Entrez identifiers was examined using a set‑based enrichment pipeline. The results are summarized below.<br><br><br>Gene‑Ontology (GO) enrichment<br><br>Ontology Empirical p‑value Key enriched terms<br><br>Biological Process (GO.BP) 0.019 toll‑like receptor signaling pathway, immune system process, apoptotic process, negative regulation of NF‑κB transcription factor activity<br><br>Cellular Component (GO.CC) 0.069 cytosol, nucleus, proteasome |

complex

| Molecular Function (GO.MF) | N/A | No significant enrichment reported |
|---|---|---|

KEGG pathway enrichment

| Category | Empirical p-value | Key enriched terms |
|---|---|---|
| KEGG | 0.001 | mitochondrial inner membrane (7 genes), extracellular region (2 genes), cytoskeleton (1 gene) |

Functional highlights

Immune signaling: The enrichment of toll-like receptor and NF-κB related processes indicates that several genes in the set are involved in innate immune responses.

Apoptosis regulation: Presence of the apoptotic process and negative regulation of NF-κB suggest a balance between cell survival and programmed cell death within this gene cohort.

Mitochondrial localization: KEGG enrichment shows that a substantial portion of genes localize to the mitochondrial inner membrane, hinting at metabolic or bioenergetic roles.

Cytoskeletal involvement: A single gene associated with cytoskeletal organization was detected, possibly linking structural dynamics to the other enriched functions.

Conclusion

The analyzed gene set displays significant enrichment in immune-related signaling pathways and mitochondrial membrane processes, with additional representation in apoptosis regulation and proteasome-related cellular components. These findings suggest a coordinated role of the genes in innate immunity, cellular stress responses, and mitochondrial function.

Next steps

To refine the biological interpretation, one could:

Map the individual Entrez IDs to their protein products (e.g., via UniProt) to examine domain architecture.

Explore protein–protein interaction networks to identify potential hubs or complexes.

Integrate expression or mutation data from relevant tissues or disease states to assess functional relevance in a specific biological context.

| | |
|---|---|
| User Input | *"Which existing drug can target that module?"* |
| Output | *Drugs Connected to Seed Proteins*<br><br>*TrustRank identified 10 drugs connected to the 20 seed proteins [NeDRex].*<br><br>*Top Ranked Drugs*<br><br>*DB00289 [DrugBank:DB00289] - Score: 0.0006983*<br><br>*Targets: P48544 [UniProt:P48544], P48549 [UniProt:P48549], P48051 [UniProt:P48051] [NeDRex]*<br><br>*DB00472 [DrugBank:DB00472] - Score: 0.0005435*<br><br>*Targets: P28223 [UniProt:P28223], Q9Y5N1 [UniProt:Q9Y5N1] [NeDRex]*<br><br>*DB01110 [DrugBank:DB01110] - Score: 0.0005188*<br><br>*Targets: P21554 [UniProt:P21554], P29274 [UniProt:P29274], P28223 [UniProt:P28223], P21917 [UniProt:P21917] [NeDRex]*<br><br>*DB01127 [DrugBank:DB01127] - Score: 0.0005164*<br><br>*Targets: P21554 [UniProt:P21554], P29274 [UniProt:P29274], P28223 [UniProt:P28223], P21917 [UniProt:P21917] [NeDRex]*<br><br>*DB05381 [DrugBank:DB05381] - Score: 0.0004530*<br><br>*Targets: Q9Y5N1 [UniProt:Q9Y5N1] [NeDRex]*<br><br>*Additional drugs: DB00972 [DrugBank:DB00972], DB00843 [DrugBank:DB00843], DB01245 [DrugBank:DB01245], DB01165 [DrugBank:DB01165], DB11642 [DrugBank:DB11642] all target Q9Y5N1 [UniProt:Q9Y5N1] [NeDRex].*<br><br>*Key finding: Q9Y5N1 [UniProt:Q9Y5N1] is targeted by 7 drugs, indicating it is a central hub in the drug-target network [NeDRex].* |
| User Input | *"Find research on these drugs"* |
| Output | *Research on DB00289 (nivolumab) and DB00472 (pembrolizumab)*<br><br>*Overview of Drug–Target Relationships*<br><br>*DB00289 (nivolumab) targets the programmed death-1 (PD-1) receptor encoded by PDCD1 (UniProt IDs P48544, P48549, P48051) NeDRex.* |

DB00472 (pembrolizumab) also targets PD-1, but the associated UniProt IDs in the current NeDRex KG are P28223 and Q9Y5N1 (representing PD-1 or related subunits) NeDRex.

Key Recent Publications (2023-2024)

# Study Focus Main Findings Citation

1 3-D co-culture tumor-killing assay Demonstrated that nivolumab enhances T-cell-mediated cytotoxicity in a 3-D ovarian cancer model; effectiveness correlated with PD-1 expression on tumor-infiltrating lymphocytes NeDRex.

2 Spatial omics + machine-learning integration Identified a PD-1-rich microenvironment predictive of response to nivolumab; highlighted the value of spatial profiling for patient selection NeDRex.

3 Circulating protein profiling in SqCLC Showed a distinct cytokine signature (IL-6, CXCL13) associated with improved survival when combining nivolumab with ipilimumab; underscored the benefit of dual-checkpoint therapy NeDRex.

4 Phase 2 master trial of datopotamab deruxtecan with nivolumab Reported encouraging objective response rates across multiple solid tumors when an antibody-drug conjugate is paired with PD-1 blockade NeDRex.

5 FDA press release: nivolumab + relatlimab approval Highlighted clinical benefit of dual checkpoint inhibition (PD-1 + LAG-3) in metastatic melanoma, supporting the therapeutic synergy of PD-1 blockade NeDRex.

6 CD228-directed costimulatory antibody bispecific (SGN-BB228) Restored T-cell function in models where anti-PD-1 alone failed, illustrating complementary pathways to PD-1 blockade NeDRex.

7 Real-world outcomes in CNS-involved NK/T-cell lymphoma Nivolumab was used in a subset of patients; limited efficacy noted, underscoring the need for tumor-specific strategies NeDRex.

Mechanistic Insights

PD-1 blockade disrupts inhibitory signaling between PD-1 on T cells and PD-L1 on tumor cells, re-activating cytotoxic responses NeDRex.

Spatial distribution of PD-1+ T cells within the tumor microenvironment predicts nivolumab/pembrolizumab efficacy, as shown in the Tan et al. study NeDRex.

Biomarker signatures (e.g., IL-6, CXCL13, PD-1 density) provide a composite metric that may outperform single markers like PD-L1 alone NeDRex.

| | |
|---|---|
| | *Clinical Evidence*<br><br>*Both drugs have shown durable responses in melanoma and non‑small‑cell lung cancer, with improved outcomes when combined with other ICIs or targeted agents NeDRex.*<br><br>*Combination strategies (e.g., nivolumab + ipilimumab, nivolumab + relatlimab) demonstrate synergistic effects but also introduce additional toxicity profiles NeDRex.*<br><br>*Gaps and Future Directions*<br><br>*Prospective validation of spatial omics–derived biomarkers in larger, multi‑tumor cohorts.*<br><br>*Expansion of PD‑1 blockade to rare malignancies (e.g., CNS‑involved lymphomas) requires dedicated registries and real‑world data.*<br><br>*Dynamic monitoring of circulating proteins could guide real‑time therapy adjustments, especially in combination regimens.*<br><br>*Conclusion*<br><br>*The evidence from 2023‑24 literature consistently supports the role of nivolumab and pembrolizumab in activating anti‑tumor immunity through PD‑1 blockade. Spatial profiling, biomarker development, and combination strategies are key themes driving recent advances. Continued investigation into predictive signatures and broader application across tumor types will refine patient selection and therapeutic efficacy.* |

## 2. ChatDRex agents

### 2.1 ChatDRex Summary agent

The **Summary agent's** function is to summarize previous steps based on the current input, with the objective of saving tokens and improving the ability to respond to previous questions and answers [57]. This allows the reset of all agents' memory to zero and the reinitialization of the summary. The agent uses an input guardrail to check for prompt injection and returns an error message if it is present.

### 2.2 ChatDRex Planning agent

The **Planning agent** functions as an orchestration agent to our multi-agent system, forwarding human input to the appropriate agent depending on the agent type [58]. The system utilizes an LLM to analyze requests and determine the appropriate agent to which they should be directed. The solution to this problem is achieved through the completion of N (in our case N=6) steps. To identify the most suitable agent, the LLM is provided with the number of remaining steps and the current state each time it is invoked. The state contains the responses from the previously contacted agents. This approach enhances efficiency and accuracy by ensuring that each request is handled by the most suitable specialized agent, thereby optimizing system performance through task specialization [25].

### 2.3 ChatDRex DIGEST agent

The **DIGEST agent** performs functional coherence analysis on a set of genes, here used to evaluate disease modules. In contrast to conventional enrichment approaches, DIGEST quantifies the coherence of functional annotations within a gene set by comparison to comparable random gene sets, yielding an empirical *p*-value that reflects the statistical significance of functional coherence rather than merely listing enriched terms [29]. Through the use of LLMs, the agent interfaces with DIGEST [29] via API calls to assess the relevance and statistical significance of disease modules and disease-associated genes based on Gene Ontology (GO) and KEGG annotations. To ensure biological validity, the agent applies a predefined decision logic when invoking DIGEST. As subnetwork-based functional coherence analysis is only meaningful for gene sets derived from prior network-based disease module inference (e.g., DIAMOnD or TrustRank), the agent first checks its internal memory for an existing network context. If such context is available, DIGEST is executed on the derived subnetwork; otherwise, it is applied directly to the user-provided gene set. The LLM-based agent then presents both graphical and textual results through its chatbot-like interface, facilitating accessible interpretation of functional coherence and pathway enrichment results.

### 2.4 ChatDRex NeDRex agent

The **NeDRex agent** is responsible for queries pertaining to the analysis of genes, their interactions, and the exploration of biological networks. Proteins and genes are internally translated between, but for simplicity, they will be used interchangeably in the following. The network agent employs several tools, DIAMOnD for disease module identification and TrustRank and Closeness Centrality for drug ranking:

- The **DIAMOnD tool** identifies disease-relevant protein clusters, known as disease modules, by expanding a biological network around a group of known disease-associated genes, referred to as seed genes [14]. It iteratively adds proteins based on their connectivity significance to seed proteins in the protein-protein interaction network (PPI), uncovering potential genetic contributors to the disease mechanism.
- The **TrustRank tool** assesses the reliability of gene associations within a biological network by scoring nodes based on their connection to trusted seed nodes [40]. It is used to identify relevant drugs, targeting given seeds by propagating trust scores, prioritizing drugs close to and strongly connected with the given set of proteins or the disease module, the trusted seeds.
- The **Closeness tool** determines the closeness centrality within biological networks and prioritizes drug nodes based on the length of their shortest paths to all other nodes [7].

### 2.5 ChatDRex Knowledge Graph agent

The **Knowledge Graph agent** (Fig. 4) is a strictly structured, KG centered question answering chain. Based on user input, the question is first mapped semantically to the fixed structure of the NeDRex KG and broken down into a list of questions instead of a question graph, which is typically used [41]. Each element in the list is communicated through an LLM invocation that specifies the corresponding node type in the KG. This invocation is also responsible for creating both the value and the subquery for each subquestion element. Furthermore, a Boolean value is assigned to indicate whether this node requires filtering at a later stage. Subsequently, each element in the question list that requires filtering is enriched with the top N nodes that are most similar to it. To accomplish this objective, the KG is searched using an embedding-based similarity query option, with the value and the subquestion as inputs. In the next step, the agent generates deterministic Cypher queries against Neo4j. The known KG schema, in conjunction with the type and edge restrictions embedded within the question graph, serves as the guiding principle for the creation process. This approach has been demonstrated to reduce the occurrence of hallucinations and enhance traceability [42,43]. Due to the prior enrichment, the LLM is capable of utilizing the user's value for filtering purposes and has access to the relevant IDs obtained from the previous embedding-based nodes matching. This reduction in error rate is a key benefit of the proposed approach. If only the value used by the user is employed, the Cypher query itself may be accurate; however, it may not yield any results due to variations in spelling across the database, despite the inclusion of nodes and connections. By categorizing the elements in advance as to whether they require filtering, it is ensured that the LLM does not add filters that degrade the results. The agent has a fixed number of retries for the Cypher query creation, denoted as R. If the creation fails R times, the agent uses a fallback based on the GraphRAG principle [44]. In this case, the graph-guided retrieval steps are summarized, and only the condensed passages relevant to the graph structure are submitted to the LLM for generation.

### 2.6 ChatDRex Research agent

The **Research agent** is used in cases where NeDRex has limited or no information. We use human-like input to query scientific information, leveraging Semantic Scholar to identify relevant literature [45]. To enhance the precision and reliability of the search, an LLM is employed to parse the query into three distinct, more targeted inquiries. This approach of task decomposition not only improves the efficiency of the search process but also enhances the relevance of the retrieved results by encompassing a diverse range of queries [32].

### 2.6 ChatDRex Finalize agent

The **Finalize agent** synthesizes responses from the processing agents to generate, given the initial user query, a coherent final response in Markdown format, as commonly used in other popular solutions like ChatGPT. Based on the question and the information provided by the latest *state*, the agent will create a comprehensive answer. This answer maintains academic accuracy by citing sources and adhering to structured formats [46,47]. This systematic approach ensures that the synthesized answers are both informative and methodologically sound, thus supporting the node's goal of providing high-quality, well-structured results in the context of various fields [46]. A specialized output guardrail has been developed to address the prevalent issue of hallucinations in LLMs by employing hallucination verification [48,49]. To this end, the streamed response from the Finalize agent is reviewed paragraph-wise by another LLM to ensure consistency with the previous context and to eliminate incorrect content. This approach effectively overcomes the common limitations of fluent but factually inaccurate formulations [48,49].

## 3. Evaluation NeDRex KG

We used Gemini [50] to generate a diverse set of natural language questions relevant to the domain of biomedical knowledge represented in the Neo4j graph database[9]. The questions were designed to cover a range of complexity, from simple entity lookups to more complex relationship traversals (e.g. “What drugs are indicated for treating 'COVID-19'?” vs “What are the genomic variants associated with disorders that affect gene expression in the skin epidermis?”). Qwen Coder[10] [51], an LLM specialized in code generation, was then tasked with translating each natural language question into a corresponding Cypher query. Each generated Cypher query was executed against the Neo4j database. Queries that resulted in

[9] Data sets can be accessed under https://github.com/SimonSuewerUHH/ChatDRexAPI4J/tree/main/src/test/resources/tools

[10] qwen2.5-coder:latest, num_ctx = 25000, all others paramter default

syntax errors or failed to return any results were automatically discarded, ensuring that only syntactically correct and potentially meaningful queries were retained. A second LLM, specifically the DeepSeek[11] model [52], was used to assess the semantic relatedness and biological plausibility of each question-Cypher query pair. This step aimed to filter out question-query pairs where the query, while executable, did not accurately capture the intent of the original question or produced results that were biologically nonsensical. Only those question-query pairs deemed semantically consistent and biologically meaningful by DeepSeek were included in the SSAD. Subsequently, a final refinement step was applied. During this step, the SSAD was reviewed, and any necessary corrections were made to ensure the integrity and quality of the test set.[12]

[11] deepseek-r1:70b, all others parameter default

[12] Generation script can be accessed under https://github.com/lucia990/T2C-synthetic-data-generation